\documentclass[dvipsnames,pbalance,authorversion,nonacm,format=sigconf,anonymous=false,review=false]{acmart}

\usepackage{hyperref}
\usepackage[inline]{enumitem}
\usepackage[nameinlink]{cleveref}
\usepackage{glossaries}
\usepackage[dvipsnames]{xcolor}
\usepackage{caption}
\usepackage{subcaption}
\usepackage{multirow}

\newcommand{\maxPapersPerYear}{21}
\newcommand{\minPapersPerYear}{12}
\newcommand{\includedPapers}{168}

\newcommand{\bbsrPapers}{18}

\newcommand{\averageCompleteness}{0.62}

\newcommand{\medianCompleteness}{0.61}
\newcommand{\paperCompletenessDecreaseYear}{2020}
\newcommand{\paperCompletenessDecreaseMedian}{0.57}
\newcommand{\paperCompletenessMaxYear}{2022 and 2024}
\newcommand{\paperCompletenessMaxMedian}{0.66}

\newcommand{\somethingAvailable}{62}
\newcommand{\somethingAvailablePerc}{36.90\%} 
\newcommand{\somethingAvailableMin}{17.64\%}
\newcommand{\somethingAvailableMax}{52.63\%}

\newcommand{\overallPaperBothCodeAndData}{10.71\%}
\newcommand{\maxPaperBothCodeAndData}{25.00\%}
\newcommand{\overallPDF}{7.14\%}
\newcommand{\maxPDF}{21.43\%}

\newcommand{\overallLinkNotWorking}{6.55\%}
\newcommand{\overallPersistent}{14}

\newcommand{\cohensKappa}{0.67}

\newacronym{aaai}{AAAI}{Association for the Advancement of Artificial Intelligence}
\newacronym{acm}{ACM}{Association for Computing Machinery}
\newacronym{ai}{AI}{Artificial Intelligence}
\newacronym{bbsr}{BBSR}{Benchmarking, Benchmarks, Software, and Reproducibility}
\newacronym{bpmn}{BPMN}{Business Process Model and Notation}
\newacronym{ec}{EC}{Evolutionary Computation}
\newacronym{ecom}{ECOM}{Evolutionary Combinatorial Optimization and Metaheuristics}
\newacronym{gecco}{GECCO}{Genetic and Evolutionary Computation Conference}
\newacronym{ijcai}{IJCAI}{International Joint Conferences on Artificial Intelligence} 
\newacronym{llm}{LLM}{Large Language Model}
\newacronym{ml}{ML}{Machine Learning}
\newacronym{neurips}{NeurIPS}{Neural Information Processing Systems}
\newacronym{rq}{RQ}{Research Question}

\begin{document}

\title{Assessing Reproducibility in Evolutionary Computation}
\subtitle{A Case Study using Human- and LLM-based Assessment}

\author{Francesca Da Ros}
\orcid{0000-0001-7026-4165}
\affiliation{%
  \institution{DPIA, University of Udine}
  \city{Udine}
  \country{Italy}}
\email{francesca.daros@uniud.it}

\author{Tarik Začiragić}
\orcid{0009-0002-9630-9354}
\affiliation{%
  \institution{LIACS, Leiden University}
  \city{Leiden}
  \country{Netherlands}}
\email{tarikzacir@gmail.com}

\author{Aske Plaat}
\orcid{0000-0001-7202-3322}
\affiliation{%
  \institution{LIACS, Leiden University}
  \city{Leiden}
  \country{Netherlands}}
\email{a.plaat@liacs.leidenuniv.nl}

\author{Thomas B\"ack}
\orcid{0000-0001-6768-1478}
\affiliation{%
  \institution{LIACS, Leiden University}
  \city{Leiden}
  \country{Netherlands}}
\email{t.h.w.baeck@liacs.leidenuniv.nl}

\author{Niki van Stein}
\orcid{0000-0002-0013-7969}
\affiliation{%
  \institution{LIACS, Leiden University}
  \city{Leiden}
  \country{Netherlands}}
\email{n.van.stein@liacs.leidenuniv.nl}

\begin{abstract}

Reproducibility is an important requirement in evolutionary computation, where results largely depend on computational experiments. In practice, reproducibility relies on how algorithms, experimental protocols, and artifacts are documented and shared. Despite growing awareness, there is still limited empirical evidence on the actual reproducibility levels of published work in the field.
In this paper, we study the reproducibility practices in papers published in the Evolutionary Combinatorial Optimization and Metaheuristics track of the Genetic and Evolutionary Computation Conference over a ten-year period. We introduce a structured reproducibility checklist and apply it through a systematic manual assessment of the selected corpus. In addition, we propose RECAP (REproducibility Checklist Automation Pipeline), an LLM-based system that automatically evaluates reproducibility signals from paper text and associated code repositories.
Our analysis shows that papers achieve an average completeness score of \averageCompleteness, and that \somethingAvailablePerc{} of them provide additional material beyond the manuscript itself. We demonstrate that automated assessment is feasible: RECAP achieves substantial agreement with human evaluators (Cohen’s $\kappa$ of \cohensKappa{}). Together, these results highlight persistent gaps in reproducibility reporting and suggest that automated tools can effectively support large-scale, systematic monitoring of reproducibility practices.
\end{abstract}

\begin{CCSXML}
<ccs2012>
    <concept>
       <concept_id>10002944.10011123.10010912</concept_id>
       <concept_desc>General and reference~Empirical studies</concept_desc>
       <concept_significance>300</concept_significance>
       </concept>
   <concept>
       <concept_id>10010405.10010497.10010498</concept_id>
       <concept_desc>Applied computing~Document searching</concept_desc>
       <concept_significance>300</concept_significance>
       </concept>
   <concept>
       <concept_id>10003752.10010070.10011796</concept_id>
       <concept_desc>Theory of computation~Theory of randomized search heuristics</concept_desc>
       <concept_significance>300</concept_significance>
       </concept>
 </ccs2012>
\end{CCSXML}

\ccsdesc[300]{General and reference~Empirical studies}
\ccsdesc[300]{Applied computing~Document searching}
\ccsdesc[300]{Theory of computation~Theory of randomized search heuristics}

\keywords{Large Language Models, Benchmarking, Scientific Reporting, Open Science, Scholar Publications}

\maketitle
\glsresetall

\section{Introduction}
\label{sec:introduction} 

Reproducibility is an important requirement of empirical research. This is especially true in \gls{ec}, \gls{ml}, and \gls{ai}, where conclusions are mainly supported by aggregated results of computational experiments. In these fields, results depend on multiple factors, including algorithm implementations, parameter settings, benchmark instances, and computing environments.
When such details are missing, ambiguous, or inaccessible, experiments cannot be reliably rerun, and reported conclusions cannot be independently verified. 

In this work, we adopt a pragmatic notion of reproducibility, focused on whether published information and artifacts are sufficient to make rerunning experiments feasible in practice. We do not assess the correctness of reported results, nor do we attempt full independent replication, but instead evaluate the presence, accessibility, and clarity of reproducibility-relevant signals.

Although reproducibility is widely recognized as important, empirical evidence on how well it is achieved in practice within \gls{ec} remains limited. 
Existing guidelines describe good experimental practices \cite{DBLP:journals/corr/abs-2007-03488,kononova2025benchmarkingmattersrethinkingbenchmarking}, but they rarely specify which information is strictly required for an experiment to be reproducible. Reviewers also lack concrete tools to assess reproducibility consistently.
As a result, reproducibility is often assumed rather than explicitly evaluated.

This paper studies reproducibility practices in \gls{ec} through a large-scale empirical analysis. We focus on full papers published in the \gls{ecom} track of the \gls{gecco} over a ten-year period. This track is one of the core and longest-running \gls{gecco} tracks and primarily publishes empirical studies of combinatorial optimization algorithms. 
Its papers typically involve stochastic algorithms, multi-factor experimental designs, and extensive empirical comparisons, all of which place particularly strong demands on transparent reporting of experimental protocols and artifacts. This makes the track a suitable stress test for reproducibility practices in empirical evolutionary computation.

Our study pursues the following goals:
\begin{enumerate*}[label=(\roman*)]
\item to assess the completeness of reproducibility reporting of \gls{ecom} publications;
\item to identify which reproducibility checklist items are consistently reported and which are systematically omitted;
\item to analyze how reproducibility artifacts are provided, maintained, and accessed over time; and
\item to evaluate the extent to which automated methods can approximate manual assessments of reproducibility signals.
\end{enumerate*}

To address these goals, we introduce a structured reproducibility checklist tailored to \gls{ec} experiments and based on \gls{acm} reproducibility standards \cite{acm-badge-guidelines}. 
We apply this checklist to \includedPapers{} \gls{ecom} papers published between 2016 and 2025, combining a manual evaluation protocol with an automated pipeline based on a \gls{llm}, called RECAP.
This allows us to compare automated and human assessments and to analyze where they agree or differ. Our focus is on the availability and clarity of information and artifacts needed to rerun experiments, not on validating the reported results. 
The automated pipeline is designed as a support tool rather than a replacement for human judgment, with the goal of improving scalability and consistency of reproducibility assessment while preserving interpretability and auditability of individual decisions.

Our contributions are as follows:
\begin{enumerate*}[label=(\roman*)]
    \item A structured reproducibility checklist tailored to empirical research in \gls{ec}, spanning methodological clarity, experimental setup, results reporting and artifacts.
    \item A mixed-method assessment protocol combining manual evaluation with an automated \gls{llm}-based pipeline (RECAP), enabling direct comparison between human and automated judgments.
    \item A ten-year analysis of reproducibility practices in \includedPapers\ \gls{ecom} papers, highlighting persistent gaps and emerging trends.
\end{enumerate*}

The remainder of this paper is organized as follows.
\Cref{sec:rel-work} reviews related work. 
\Cref{sec:methodology} describes our methodology.
\Cref{sec:results} presents the results of the empirical analysis, including longitudinal trends and a comparison between manual and automated assessments.
\Cref{sec:discussion} discusses observed limitations and practical recommendations for improving reproducibility reporting.
Finally, \Cref{sec:conclusions} concludes and outlines directions for future work. 

\section{Related Work}
\label{sec:rel-work}

Reproducibility has become a central concern across scientific disciplines~\cite{baker-nature}, and particularly within computer science~\cite{DBLP:journals/cacm/CockburnDBG20,DBLP:journals/aim/DesaiAP25,DBLP:conf/aaai/RaffBSF25}, where much of the evidence underlying research claims comes from experiments. 
Large-scale assessments in \gls{ai} and \gls{ml} consistently show that documentation practices are often insufficient to enable reliable reproduction. 
For instance, \citet{DBLP:conf/aaai/GundersenK18} reviewed 400 papers from \gls{ijcai} and \gls{aaai} and concluded that none of the examined studies were fully reproducible, although modest improvements were observed over time. 
A subsequent study by \citet{DBLP:conf/aaai/GundersenCMN25} assessed 30 highly influential \gls{ai} papers and found that access to original materials substantially increases the likelihood of successful reproduction, yet divergence from reported results remains common. 
Several surveys further examine methodological sources of irreproducibility, identifying numerous design and reporting decisions that can bias conclusions or hinder reproducibility, including undocumented preprocessing steps, incomplete descriptions of experimental conditions, and missing implementation artifacts~\cite{DBLP:journals/aim/GundersenGA18,DBLP:journals/jmlr/PineauVSLBdFL21,DBLP:conf/zeus/HummelM24,DBLP:journals/corr/abs-2204-07610}.

\gls{ec} is by no means exempt from this \emph{reproducibility crisis}.
Although concerns about the reliability of empirical results in randomized search heuristics date back to the late nineties~\cite{DBLP:conf/dimacs/Johnson99,DBLP:conf/cp/BeckDF97}, the field has now reached a critical point. \citet{DBLP:journals/telo/Lopez-IbanezBP21} highlighted several challenges that \gls{ec} shares with other computer science fields, as well as others specific to the domain. 
These include the overreliance on competitive testing, biases toward publishing positive results, the scarcity of standardized benchmarks, the difficulty of reproducing empirical comparisons, and the sensitivity of stochastic optimization to implementation details~\cite{DBLP:journals/eor/TurkesSH21,DBLP:journals/eor/SwanABHJKKMMOPG22,DBLP:conf/recsys/Shehzad0MJ25,DBLP:journals/itor/Sorensen15}.

Several initiatives aim to address these issues, both in the broader \gls{ai} field and within the \gls{ec} community. 
The \gls{acm} Artifact Review and Badging process provides a unified framework for assessing the availability, completeness, and reusability of digital artifacts~\cite{acm-badge-guidelines}. 
Open science principles further encourage the publication of source code, datasets, experimental logs, and metadata, as well as open-access dissemination~\cite{open-access-networs,MILJKOVIC2024100096}. 
Major conferences have begun adopting reproducibility checklists~\cite{aaai-checklist,ijcai-2025-checklist}, dedicated reproducibility tracks~\cite{sigir-repro-track,ecir-2025-reproduc}, and tutorials on reproducible research practices~\cite{DBLP:conf/ecir/FerraraPT25}. 
Similar principles apply to \gls{ec}, where open-source libraries~\cite{DBLP:conf/gecco/MedvetNM22,DBLP:journals/spe/GasperoS03}, benchmark repositories~\cite{DBLP:journals/oms/HansenARMTB21,roarnet-problem-statements}, and guidelines for rigorous experimental design~\cite{DBLP:journals/corr/abs-2007-03488,IOHanalyzer} are increasingly common. 
At \gls{gecco}, the introduction of the \gls{bbsr} track in 2024 reflects a growing commitment to transparent and reproducible research. 
However, upon inspection of the \bbsrPapers{} papers accepted to the track, only one explicitly focuses on replication as a primary research objective~\cite{DBLP:conf/gecco/KarnsD25}, suggesting that this type of study remains comparatively underrepresented even within venues dedicated to reproducibility.

\section{Methodology}
\label{sec:methodology}

In this section, we describe the methodology adopted to systematically assess reproducibility practices in the considered papers. 
Our approach is designed to be descriptive and comparative, aiming at characterizing how reproducibility-relevant information is reported. It is out of our scope to evaluate the scientific validity or performance of the proposed methods. Thus, we adopt a pragmatic and information-centric notion of reproducibility. 
Specifically, we assess whether a paper provides sufficient information to make independent reproduction of the reported experiments \emph{in principle} possible, without requiring direct interaction with the authors. 

This perspective is aligned with established artifact evaluation frameworks, like the \gls{acm} one~\cite{acm-badge-guidelines}, which distinguish between different levels of reproducibility based on artifact availability, functionality, and reusability. 
Within this scope, we operationalize reproducibility through a structured checklist that captures the reporting of such elements in a consistent manner.

We first describe the construction of the paper corpus in \Cref{sec:methodology:corpus-construction}. 
We then introduce the reproducibility checklist developed for this study in \Cref{sec:methodology:checklist}. 
Finally, in \Cref{sec:methodology:manual-assessment,sec:methodology:llm-assessment}, we detail the protocols adopted for the human- and \gls{llm}-based assessments.

\subsection{Material Collection}
\label{sec:methodology:corpus-construction}

We retrieve all full papers published in the \gls{ecom} track between 2016 and 2025 from the \gls{acm} Digital Library~\cite{gecco-proceedings}. 
To avoid conflicts of interest during the manual assessment, we exclude any papers co-authored by authors of the present study. 

In total, our analysis considers \includedPapers{} papers. The smallest number of \gls{ecom} full papers is observed in 2021 (\minPapersPerYear{}), while the largest is in 2017 (\maxPapersPerYear{}).
Appendix~\ref{sec:material-collection-pipeline} reports a complete overview of the collection process.

\subsection{Reproducibility Checklist}
\label{sec:methodology:checklist}

We evaluate \gls{ec} reproducibility practices using a comprehensive checklist structured into 5 dimensions: 
\begin{enumerate*}[label=(\roman*)]
    \item methodological clarity,
    \item experimental setup,
    \item results reporting,
    \item artifact evaluation, and
    \item paper metadata.
\end{enumerate*}
Each dimension contains several items, and for each item we record whether the relevant information is present (\textsf{Y}/\textsf{Yes}), absent (\textsf{N}/\textsf{No}), or not applicable (\textsf{NA}). 
For a small subset of items, the annotation is performed by selecting one value from a set of admissible options (e.g., for \textsf{Parameter configuration}, one may choose between \textsf{Manual}, \textsf{Automated}, or \textsf{NA}).

The checklist spans both high-level characteristics (e.g., \textsf{Affiliations} and \textsf{Best-paper nomination}) and core reproducibility indicators (e.g., \textsf{Hardware/machine description, Tuning budget}). 

The checklist draws inspiration from existing guidelines and checklists~\cite{aaai-checklist,ijcai-2025-checklist}, but is specifically tailored to the methodological and experimental characteristics of \gls{ec} research.
The artifact-related dimension follows the \gls{acm} Artifact Review and Badging guidelines~\cite{acm-badge-guidelines}, assessing availability, documentation quality, reusability, and whether results can be reproduced or replicated. 

The complete checklist is reported in Appendix~\ref{sec:reproducibility-checklist}.

\subsection{Manual Assessment Protocol}
\label{sec:methodology:manual-assessment}

We conduct a manual assessment of each paper following a predefined protocol and using our checklist. 
The assessment was performed by inspecting the paper, the supplementary material included in the proceedings, any publicly available artifacts referenced by the authors, and the \gls{gecco} website.\footnote{Note that the supplementary material can include PDF material and/or the artifacts.}

\begin{figure}[ht]
    \centering
    \includegraphics[width=\linewidth]{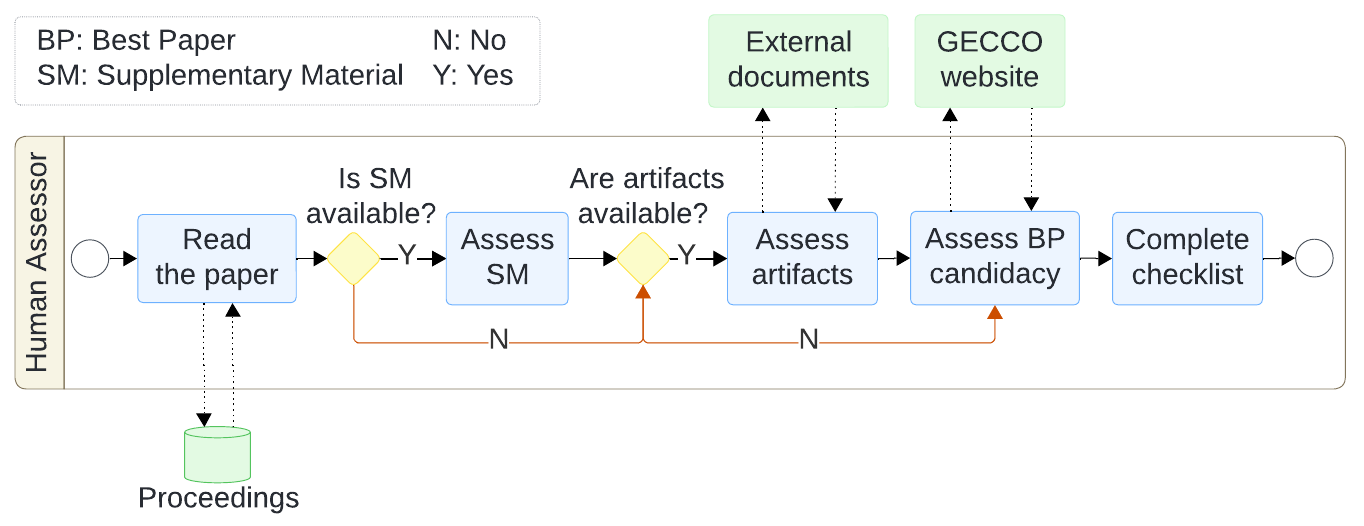}
    \caption{\gls{bpmn} model of the manual assessment of a paper.}
    \label{fig:manual-assessment}
    \Description{BPMN representation of the reproducibility assessment process adopted for ECOM papers. For each paper, a human assessor reads the manuscript, searches for supplementary material in the proceedings, and identifies referenced artifacts. When artifacts are available, they are evaluated, allowing the use of external documentation but excluding LLM assistance. Information on best paper nomination and awards is retrieved from the conference website. All outcomes are systematically recorded using the proposed checklist.}
\end{figure}

\Cref{fig:manual-assessment} reports a \gls{bpmn} model~\cite{10.5555/3035641} of the manual assessment.
We first thoroughly read the paper and look for any supplementary material on the proceeding website. Then, we identify all artifacts referenced by the authors, including source code repositories, scripts, configuration files, and datasets. 
When the code is available, we evaluate the feasibility of rerunning the experiments. Code assessment is conducted under a fixed time budget of two hours, which includes both environment setup and execution. 
During this process, the assessor is allowed to consult external online resources (e.g., documentation or dependency information) to support setup and execution, but does not use \gls{llm} assistance.
Information related to the best paper nomination and win are retrieved using the conference websites (each year \gls{gecco} relies on a new website).

All assessment outcomes are recorded using our checklist. To ensure a conservative evaluation, checklist items are labeled as \textsf{Y} whenever at least some explicit and valid evidence addressing the item is present. 
For example, if only a subset of the testing instances is made available, the checklist item \textsf{Testing instances specified} is marked as \textsf{Y}. 
Similarly, if at least a subset of the parameters is tuned using an automated pipeline, the checklist item \textsf{Parameter configuration} is labeled as \textsf{Automated}.

\subsection{RECAP: LLM-based Assessment Protocol}
\label{sec:methodology:llm-assessment}

\begin{figure}[ht]
    \centering
    \includegraphics[width=\linewidth]{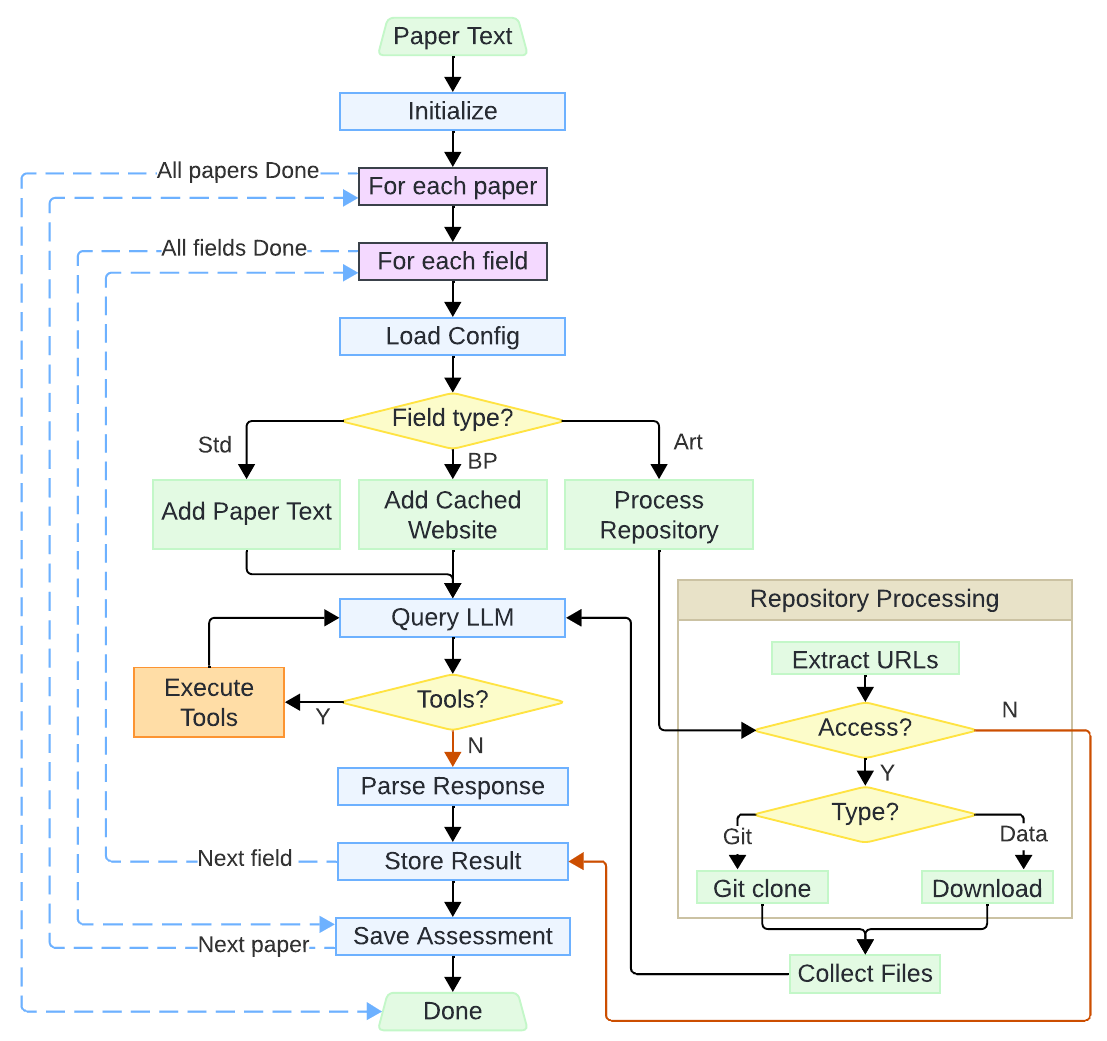}
    \caption{RECAP system overview. The system processes each paper through a field-by-field evaluation loop. Based on field type, it either uses the paper text directly (Std), retrieves cached best paper website data (BP), or processes linked repositories (Art). Each field is evaluated by an \gls{llm} with the appropriate context, and some fields invoke tools (e.g., sandbox execution). Results are parsed and stored per paper.} 
    \label{fig:RECAP_diagram}
    \Description{Pipeline describing the LLM-based assessment.}
\end{figure}

In order to help both paper authors and conference organizers uphold a standard of reproducibility, we devise the \textbf{RECAP} (REproducibility Checklist Automation Protocol) system. RECAP uses our checklist as a base. Each field from the checklist is described in a concise and structured manner with specific criteria for what constitutes each of the possible answers.\footnote{Part of the authors worked on RECAP and part on the Manual Assessment Protocol to eliminate any possibility of introducing bias into RECAP by ways of describing what serves as a \textsf{Y}, what as a \textsf{N}, etc.}
The evaluation process for each field is encapsulated in a system prompt that guides the \gls{llm} to answer the question. For this paper, we use OpenAIs GPT 5 nano model, because of its cost/performance trade-off~\cite{openai2025gpt5nano}. 
Importantly, in our system the \gls{llm} is used primarily as an information retrieval and entity extraction tool rather than being assigned a reasoning role. We deliberately limit reasoning tasks for the \gls{llm} to prevent problems with hallucinations and confusion that can arise when \glspl{llm} are asked to make complex judgments.
\Cref{fig:RECAP_diagram} shows a mid-level flowchart of how the system operates. For all \includedPapers{} papers, we convert the PDF files to plain text, extracting only the textual content.\footnote{Figures and tables are not extracted, as we believe a well-written paper will have sufficient captions explaining what each figure conveys.} The system processes each paper by iterating through all checklist fields. For each field, we construct an \gls{llm} context containing a customized system prompt, a structured JSON response schema, and the full paper text. Due to the large context windows of modern \glspl{llm}, we send the entire paper text in a single request rather than relying on chunking or retrieval-augmented approaches. Depending on the field type, additional context may be appended.
For standard fields, such as those evaluating methodology descriptions, experimental setup details, or statistical reporting, the \gls{llm} returns a structured response indicating whether the criterion is met (\textsf{Y}), not met (\textsf{N}), or not applicable (\textsf{NA}), along with a brief disambiguation.\footnote{This disambiguation does not properly serve as explainability due to how \glspl{llm} generate responses, but serves as a proxy to gauge whether errors occurred. The answer and disambiguation should be treated as separate entities.}
For artifact-related fields, the system extracts repository URLs from the paper, then clones \textsf{Git} repositories or downloads data files into an isolated \textsf{Docker} sandbox pre-configured with common scientific dependencies. The truncated contents of all text-based (code/data) files are presented to the \gls{llm} for assessing documentation quality, code completeness, and consistency with the paper's methodology. For functional exercisability specifically, the system attempts to run the code. Due to computational constraints, execution is limited to five minutes, receiving \textsf{Y} if successful, \textsf{N} if errors occur.
All responses conform to predefined JSON schemas, and each paper's complete assessment is saved as an individual JSON file for subsequent analysis. 

\section{Results}
\label{sec:results}

We summarize the outcomes of the manual assessment in \Cref{sec:results:analytical-results}, providing an overview of reproducibility practices in the \gls{ecom} track of \gls{gecco} over the years.
We then report, in \Cref{sec:results:manula-vs-automated}, a comparative analysis between the results obtained with RECAP and those derived from the manual assessment.
Appendix~\ref{sec:metrics-definition} describes the metrics we use.

\subsection{Reproducibility Assessment}
\label{sec:results:analytical-results}

We address the following \glspl{rq}, relying exclusively on the manually compiled checklists.
\begin{enumerate}[label = \textbf{RQ}\arabic*,topsep=0.5em,leftmargin=1em,labelwidth=*,align=left]
    \item How does the overall reproducibility completeness of \gls{ecom} papers vary across years? (\Cref{sec:results:analysis:paper-level}).
    \item Which reproducibility items are most consistently reported, and which are systematically omitted? (\Cref{sec:results:analytical:item}).
    \item How do \gls{ecom} papers provide reproducibility artifacts, and how stable are these practices? (\Cref{sec:results:analytical:artifacts}).
    \item Do the best paper candidates exhibit different reproducibility reporting patterns compared to other papers? (\Cref{sec:results:analytical:best})
\end{enumerate}

\subsubsection{Paper-level analysis}
\label{sec:results:analysis:paper-level}

This section analyzes the evolution of paper-level reproducibility completeness in the \gls{ecom} track over time. 
The goal is to quantify how consistently reproducibility-related information is reported across years.

For each paper, we define its reproducibility completeness as the proportion of checklist items answered positively, excluding those marked as \textsf{NA}. The exclusion of \textsf{NA} items ensures that papers are not penalized for aspects that are outside their scope. 
For example, for papers of type \textsf{Theory}, it is not meaningful to assess experimental-related items. 
Similarly, when a paper does not employ any statistical test (answer \textsf{N} to \textsf{Statistical test}), items concerning the reporting of statistical significance levels are not applicable and are therefore excluded from the computation. 
Furthermore, checklist items that convey descriptive, rather than the presence or absence of reproducibility-related content, are excluded from the computation (e.g., \textsf{Parameter configuration}).

\begin{figure}[ht]
    \centering
    \includegraphics[width=\linewidth]{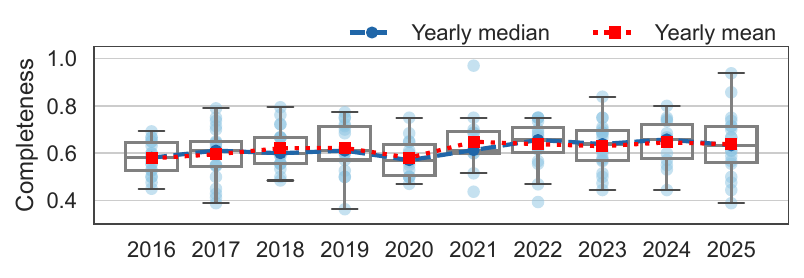}
    \caption{Paper-level completeness of reproducibility reporting across years. Individual papers are shown as points, while boxplots indicate quartiles and lines identify yearly trends.}
    \Description{Paper-level completeness of reproducibility reporting across years. Empty boxplots and jittered points show the distribution of completeness scores for individual papers. Lines show medians and means per year.}
    \label{fig:completeness}
\end{figure}

\Cref{fig:completeness} summarizes the evolution of the completeness of the paper over the years. We emphasize that this analysis captures changes in reporting practices rather than any experimental quality.
The median completeness across all papers is \medianCompleteness, with a temporary decrease around \paperCompletenessDecreaseYear{} (median of \paperCompletenessDecreaseMedian) followed by a consistent increase in subsequent years (up to a median of \paperCompletenessMaxMedian{} in  \paperCompletenessMaxYear).
At the same time, substantial dispersion persists across all years, indicating heterogeneous reporting practices even in recent editions.

We apply a Kruskal–Wallis non-parametric test with significance level $\alpha = 0.05$ to assess whether statistically significant differences exist between the groups. The test evaluates the null hypothesis that all groups are drawn from the same distribution. The resulting $p$-value of $0.36$ does not provide evidence to reject the null hypothesis.

\subsubsection{Per-item analysis}
\label{sec:results:analytical:item}

This section examines reproducibility reporting at the level of individual checklist items, so to understand which aspects contribute most to missing information.

\begin{figure}
    \centering
    \includegraphics[width=\linewidth]{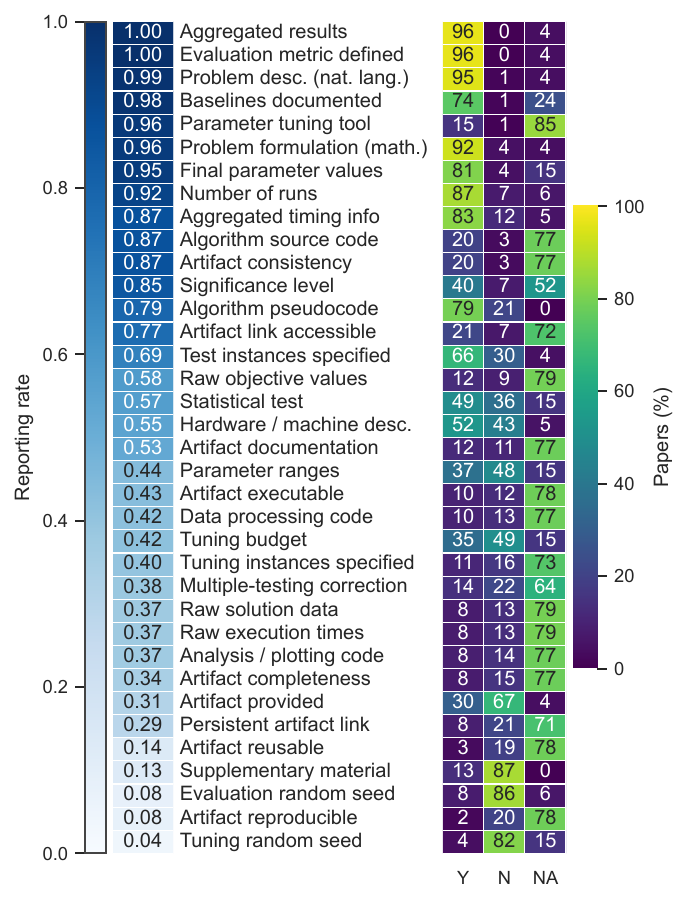}
    \caption{Reporting rate (left) per item and outcomes (\textsf{Y}, \textsf{N}, and \textsf{NA}) across papers (right).}
    \label{fig:per-item}
    \Description{}
\end{figure}

\Cref{fig:per-item} provides a per-item view of reproducibility reporting practices. 
Each row corresponds to a single checklist item. The right-hand heatmap reports, for each item, the proportion of papers in which the information is reported (\textsf{Y}), omitted (\textsf{N}), or not applicable (\textsf{NA}), while the left panel summarizes this information through an applicability-aware reporting rate, computed as the fraction of \textsf{Yes} outcomes among applicable cases (\textsf{Yes} $+$ \textsf{No}).

There is a structural imbalance in reproducibility practices: on the one hand, descriptive items (such as \textsf{Aggregated results}) are consistently documented; on the other, operational details (such as \textsf{Parameter ranges} in the manual or automated tuning) are reported far less reliably. This pattern points to community-level reporting norms rather than isolated omissions.
The least reported aspects concern artifacts and the specification of random seeds required for replication.
Beyond overall completeness, only 3 papers out of \includedPapers{} (1.79\%) provide sufficient material to attempt a full reproduction of the results without directly contacting the authors.\footnote{Note that this notion is aligned with the \gls{acm} artifact evaluation and badging framework, which defines a hierarchy of artifact levels, ranging from availability, through functionality, to reproducibility.}

This analysis also reveals additional aspects, most notably practices related to parameter tuning and statistical analysis.
With respect to parameter tuning, only 25 of the \includedPapers{} (approximately 15\%) employ automated tuning procedures, indicating that manual or fixed parameter settings remain the dominant practice.
Concerning statistical analysis, although statistical tests are reported in nearly half of the papers (48.81\%), the adoption of multiple test correction methods is rare (13.69\%).

\subsubsection{Artifact analysis}
\label{sec:results:analytical:artifacts}

This section investigates how papers in the \gls{ecom} track provide reproducibility artifacts and how stable these practices appear over time.
In the context of \gls{gecco}, authors may make reproducibility artifacts available either by including supplementary material within the proceedings or by providing externally hosted artifacts via hyperlinks. 

\begin{figure}[ht]
    \centering
    \includegraphics[width=\linewidth]{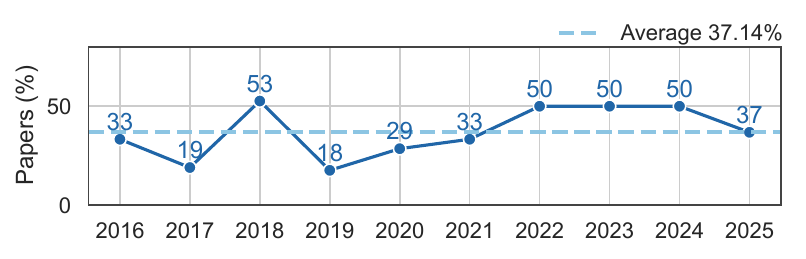}
    \caption{Proportion of papers providing some additional material other than the paper itself (supplementary material or external artifact) over time.}
    \label{fig:fig-some-material}
    \Description{The figure reports a line chart. The x-axis reports the 10-year interval, from 2016 to 2025. The y-axis reports the proportiong of paper providing either supplementary material or an external hyperlink. A dashed line report the average over the years.}
\end{figure}

Out of \includedPapers{} analysed papers, only \somethingAvailable{} (\somethingAvailablePerc) provide at least one reproducibility artifact.
\Cref{fig:fig-some-material} reports the proportion across the considered years. Artifact availability exhibits marked variability over time, ranging from values below \somethingAvailableMin{} to peaks of around \somethingAvailableMax. 
This highlights heterogeneous artifact-provision practices within the \gls{ecom} track.

\begin{figure}[ht]
    \centering
    \begin{subfigure}{\linewidth}
    \centering
    \includegraphics[width=\linewidth]{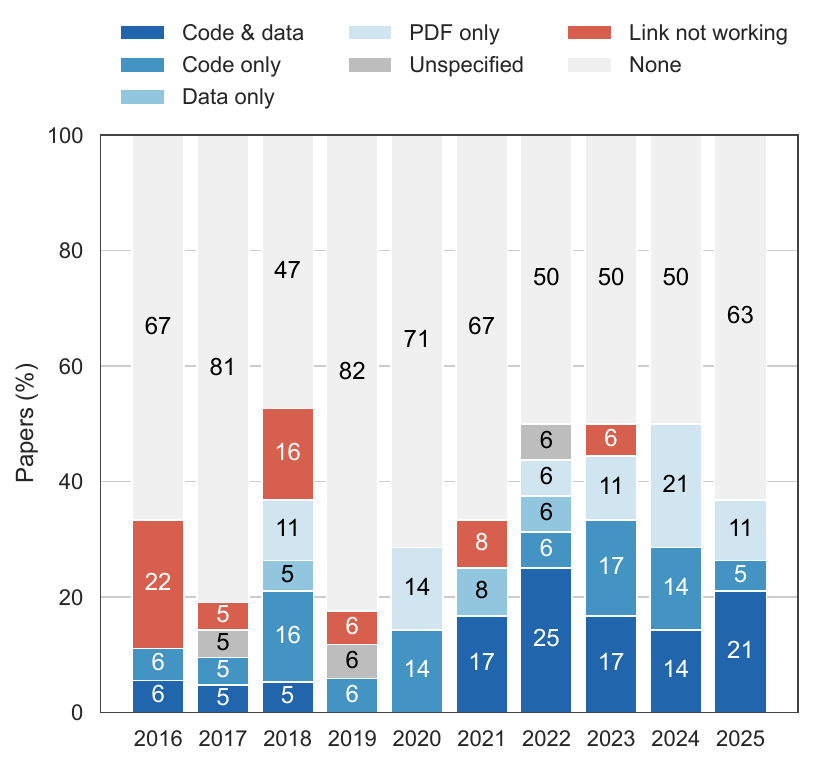}
    \label{fig:modality:year}
    \caption{Year-wise distribution.}
    \Description{Stacked barchart reporting for each year (x-axis), the percentage of papers by modality (y-axis.)}
    \end{subfigure}
    \begin{subfigure}{\linewidth}
    \centering
    \includegraphics[width=\linewidth]{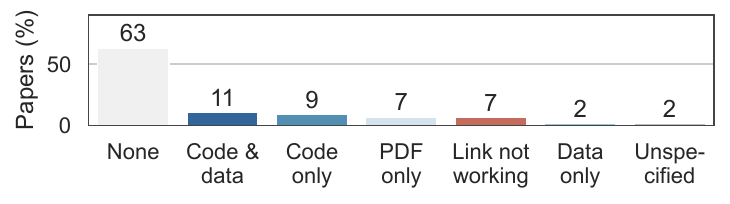}
    \caption{Overall.}
     \label{fig:modality:overall}
    \Description{Barchart reporting, for each modality (x-axis), the percentage of papers (y-axis).}
    \end{subfigure}
    \caption{Distribution of reproducibility artifact modalities.}
    \label{fig:modality}
\end{figure}

We next analyze the types of reproducibility artifacts provided, distinguishing between different artifact modalities. 
Specifically, we consider: PDF-only supplementary material; artifacts including executable components, such as source code (i.e., algorithmic code, analysis code, or pre-/post-processing code) and/or raw experimental data (i.e., complete solutions, objective values, or execution times); and unspecified artifacts, corresponding to working links that either do not contain material directly relevant to the paper or provide only problem instances.
\Cref{fig:modality} summarises the distribution of these artifact modalities in the \gls{ecom} track from both a temporal and an aggregate perspective. 
PDF-only supplementary material accounts for a non-negligible share of artifact provision (\overallPDF{} overall, with peaks of \maxPDF{} in individual years). % \footnote{\fdr{PDF materials come only in the format of Supplementary Material.}}
From 2021 onwards, an increasing number of papers provide both code and data, reaching \overallPaperBothCodeAndData{} overall and up to \maxPaperBothCodeAndData{} in specific years.

Finally, the analysis of artifact modalities highlights an additional aspect worth discussing, namely, the stability of artifact-provision practices. In particular, a non-negligible fraction of papers (\overallLinkNotWorking) provide links that are no longer working at the time of assessment (2025-11-01 to 2026-01-10). Overall, only \overallPersistent{} papers rely on persistent dissemination mechanisms, namely archival repositories such as \textsf{Zenodo} or the supplementary material option provided by the conference (here we consider only those papers providing code and/or data). 

\subsubsection{Best paper candidates}
\label{sec:results:analytical:best}

This section investigates whether papers nominated as best paper candidates exhibit different reproducibility patterns compared to the others. 
We partition the papers into 2 groups: those nominated as best paper (26 papers) and those not nominated for the award (146 papers). Further, we also report the patterns of the papers that received the award (10 papers).\footnote{Note that the metrics on nominated papers already account for them.}

Nominated papers exhibit slightly higher average (0.64) and median completeness (0.63) compared to non-nominated papers (avg.\ 0.61, median of 0.61).
Award-winning papers do not show a higher completeness score (avg.\ 0.60, median 0.60).
To assess whether the observed differences in reporting completeness between nominated and non-nominated papers are statistically meaningful, we complement the descriptive analysis with a non-parametric comparison by applying a two-sided Mann–Whitney U test to paper-level completeness scores with $\alpha=0.05$. The test does not reveal a significant difference between the two groups ($p$-value=0.57, $U$=2121.0). The Common-Language Effect Size (CLES) is small (0.57), indicating that a randomly selected best paper candidate has a 57.44\% probability of exhibiting higher completeness than a randomly selected non-candidate. Similarly, a comparison between the non-nominated and the winners reports a $p$-value of 0.79 ($U$=674, CLES = 0.47).

A clear difference emerges when considering artifact provision, where nominated (50.00\%) and winning papers (70.00\%) have higher scores w.r.t.\ non-nominated ones (34.51\%). One possible explanation lies in the structure of the completeness score itself. E.g., a paper that provides an artifact but fails to satisfy multiple associated requirements may receive a lower completeness score than a paper that does not provide any artifact at all, since missing mandatory artifact-related fields are penalized cumulatively.

\subsection{Manual vs.\ Automated Assessment}
\label{sec:results:manula-vs-automated}

We now address \textbf{RQ5}, that is, to what extent does the RECAP system reproduce manual judgements? This analysis relies on both the automated and the manual assessments, with the latter serving as a reference annotation rather than an absolute ground truth.

Note that 13 of the 168 papers could not be processed due to our approach of loading all repository files (the first 1,000 tokens of each file) into context, which exceeded token limits for repositories with many files; a more selective file handling strategy could address this limitation in future work. A detailed list of unprocessed papers is provided in Appendix~\ref{sec:recap-paper-exclusion-list}.

\begin{figure}[ht]
    \centering
    \includegraphics[width=\columnwidth]{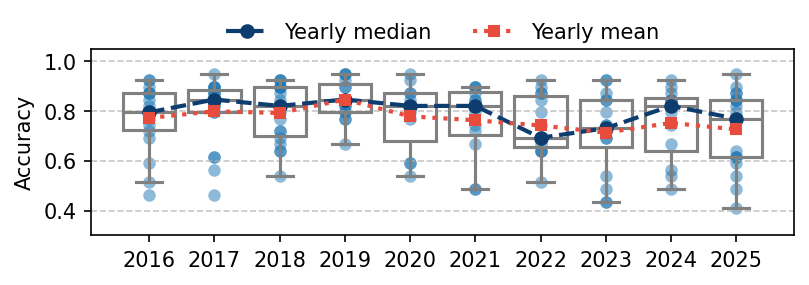}
    \caption{Distribution of per-paper prediction accuracy by publication year. Individual papers are shown as points, while box plots indicate quartiles and lines identify yearly median/mean trends.}
    \label{fig:RECAP_per_paper_accuracy}
    \Description{Distribution of per-paper prediction accuracy by publication year. Individual papers are shown as points, with box plots indicating quartiles and yearly median/mean trends.}
\end{figure}

\begin{figure}[ht]
    \centering
    \includegraphics[width=\columnwidth]{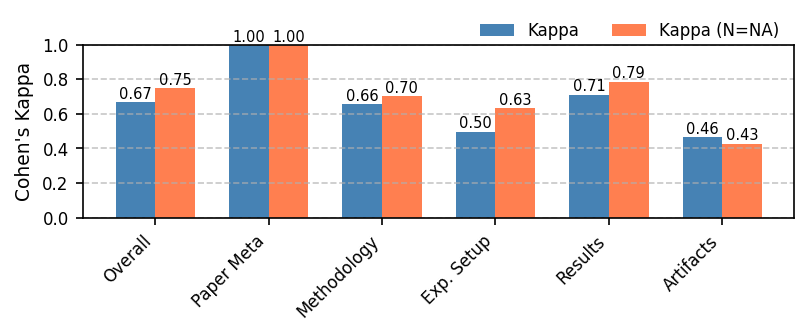}
    \caption{Inter-rater agreement (Cohen's $\kappa$ or Kappa) between RECAPs predictions and human assessment, overall and by checklist category. \textsf{Kappa (N=NA)} treats \textsf{No} and \textsf{NA} as equivalent.}
    \label{fig:RECAP_kappa_comparison}
    \Description{Inter-rater agreement (Cohen's Kappa) between RECAPs predictions and human assessment, overall and by checklist category. Kappa (N=NA) treats 'No' and 'NA' as equivalent.}
\end{figure}

We first evaluate the accuracy of the automated labeling.
\Cref{fig:RECAP_per_paper_accuracy} reports the results by year. RECAP achieves a mean per-paper accuracy of 76.8\% and a median accuracy of 79.5\% when compared against human assessment. For reference, random guessing among the three classes (\textsf{Y}, \textsf{N}, and \textsf{NA}) would result in an expected accuracy of approximately 33\%.

Since human assessment may reflect individual annotator biases, we also compute Cohen's $\kappa$ (Kappa)~\cite{landis1977kappa}, treating the human and RECAP assessments as two independent raters. As shown in \Cref{fig:RECAP_kappa_comparison}, the overall $\kappa$ of 0.67 indicates substantial agreement. Recognizing that distinguishing between \textsf{No} (not reported) and \textsf{NA} (not applicable) is inherently challenging, we also compute $\kappa$ with \textsf{N} and \textsf{NA} treated as equivalent. The resulting higher values (0.75 overall) confirm that much of the disagreement stems from this distinction rather than from fundamental classification errors.
Appendix~\ref{sec:additional-recap-analysis} includes additional analyses, 
providing deeper insight into RECAP's error patterns and highlighting areas for potential improvement. 
Overall, these results demonstrate RECAP's potential as a tool for automated reproducibility assessment. 

\section{Discussion}
\label{sec:discussion}

In this section, we first examine recurring limitations observed in the assessed papers, highlighting common pitfalls in the reporting of experimental protocols and research artifacts (\Cref{sec:discussion:papers}). We then discuss the limitations inherent to our own evaluation process (\Cref{sec:discussion:manual}) and the limitations of RECAP (\Cref{sec:discussion:automated}). Finally, we outline a set of practical recommendations aimed at improving reproducibility practices (\Cref{sec:discussion:recommendations}).

\subsection{Limitations in Assessed Papers}
\label{sec:discussion:papers}

To better contextualize the results of our analysis, we now discuss a set of recurring pitfalls observed across the assessed studies. 
Importantly, these issues should not be interpreted as shortcomings of individual authors, but rather as symptoms of broader structural and cultural practices in experimental algorithmics. 

\begin{itemize}[topsep=0.5em,leftmargin=1em,labelwidth=*,align=left]
    \item \textbf{Artifact availability and temporal fragility}: Artifact accessibility is inherently time-dependent: repositories may be updated, relocated, or removed after publication. Our analysis reflects availability at a specific timepoint (2025-11-01 to 2026-01-10) and therefore captures the \emph{current} rather than the original status of artifacts, highlighting the fragility of non-persistent hosting solutions. At the same time, although publisher-supported supplementary material offers a stable dissemination channel, it is often underutilized and primarily used for additional tables rather than for hosting structured reproducibility artifacts. Related to this, several papers state that code or data will be provided or is available upon request, yet no artifact could be retrieved at the time of assessment. Such claims may have been introduced during submission and not verified at camera-ready stage.

    \item \textbf{Ambiguity and lack of experimental specification}: Many papers describe experimental choices using informal or underspecified terminology (e.g., \emph{standard parameters, default settings, as commonly done}), making it difficult to infer exact configurations. In addition, several studies do not clearly distinguish between tunable parameters, fixed parameters, and higher-level experimental design choices, obscuring what was optimized versus what was assumed as part of the algorithmic design. Related to this, several papers report key reproducibility information across sections, footnotes, or appendices without clear signposting, increasing cognitive load and the risk of unintentional omissions.

    \item \textbf{Reliance on implicit community knowledge and external material:} Several papers rely on assumed familiarity with benchmark sets, algorithm variants, or experimental conventions. Such assumptions, while fine within the community, reduce accessibility and reproducibility for newcomers or researchers from other fields. Related to this, in some cases, essential methodological details are deferred to previous publications, technical reports, or academic theses, creating long dependency chains that weaken the self-contained nature of the paper and introduce additional barriers to independent reproduction.

    \item \textbf{Statistical analysis limitations}: Some studies report multiple statistical tests without explicitly addressing multiple-testing corrections or clearly stating the adopted significance level. This limits the robustness of reported performance differences, particularly in large-scale experimental comparisons~\cite{DBLP:journals/sigir/Fuhr17}.

    \item \textbf{Environment, licensing, and platform dependencies}: Reproducibility is further affected by missing information on software licenses, operating systems, hardware configurations, and environment setup. In several cases, code artifacts implicitly assume specific platforms or local configurations, hindering execution on different machines. 

    \item \textbf{Limited repository structuring and documentation}: Even when code artifacts are available, they often lack clear structure, documentation, or execution guides. In particular, README files with dependency specifications and minimal usage examples are frequently missing. In a few cases, repositories consist primarily of compressed archives with no evident organization, further complicating inspection and reuse.
\end{itemize}

\subsection{Limitations of our Study and of Manual Assessment}
\label{sec:discussion:manual}

We now discuss the main limitations of our study, with particular attention to our checklist and the manual assessment. 

\begin{itemize}[topsep=0.5em,leftmargin=1em,labelwidth=*,align=left]
    \item \textbf{Human bias}: Although we rely on a structured checklist, parts of the assessment necessarily involve human judgment (e.g., whether information is \emph{sufficiently detailed} or \emph{actionable}). This introduces potential subjectivity~\cite{DBLP:journals/ipm/SopranoRBCSDM24}, which we mitigate through explicit guidelines and, where applicable, cross-checking, but cannot eliminate entirely. Moreover, assessors’ familiarity with specific problem domains, benchmarks, or algorithmic conventions may influence their ability to interpret implicit or sparsely documented information.

    \item \textbf{Discretization and aggregation effects}: Most checklist items are encoded using a three-level scale (i.e., \textsf{Y, N, NA}) and subsequently aggregate into paper-level or item-level scores. This discretization simplifies what is, in practice, a graded and continuous notion of evidential quality. Many papers provide partial or fragmented information that does not fit cleanly into binary categories and is therefore absorbed into coarse decision boundaries. As a result, the aggregate scores should be interpreted as approximate indicators.

    \item \textbf{Scope of reproducibility claims}:  Our analysis reflects observable reporting practices and artifact accessibility. Consequently, our findings should be interpreted as indicators of \emph{potential reproducibility}, rather than guarantees of correctness, robustness, or empirical validity of the reported results.

    \item \textbf{Heterogeneity of research paradigms and protocols}:  The assessed papers span different experimental cultures, ranging from traditional combinatorial optimization to learning-based approaches. These paradigms differ in what is typically reported, which complicates direct comparison and may bias the assessment against papers that follow less standardized experimental conventions. Furthermore, many papers report multiple experiments within a single contribution, often involving different problem domains, algorithm variants, or experimental protocols. Our analysis assigns reproducibility levels at the paper level and can obscure differences between individual experiments.

    \item \textbf{Scalability constraints}:  Finally, manual assessment does not scale easily to very large sets of papers. While our study prioritizes depth and interpretability over scale, these constraints motivate the complementary usage of automated approaches.
\end{itemize}

\subsection{Limitations of RECAP}
\label{sec:discussion:automated}

Despite RECAP's promising results, some limitations should be acknowledged.
\begin{itemize}[topsep=0.5em,leftmargin=1em,labelwidth=*,align=left]
\item \textbf{Context window constraints}: Although modern \glspl{llm} offer large context windows, our approach of loading entire repository contents into context proved insufficient for a small portion of papers. Repositories with extensive codebases or numerous files exceeded token limits, resulting in 13 of 168 papers being unprocessable. A more selective approach, such as intelligently filtering relevant files or summarizing repository structure before detailed analysis, could mitigate this limitation.
\item \textbf{\gls{llm} hallucinations.}: Like all \gls{llm}-based systems, RECAP is susceptible to hallucinations, where the model may generate plausible-sounding but incorrect assessments. While we try to mitigate this through structured output formats and providing all relevant information as context, we cannot fully eliminate this behavior. Users should treat RECAP's outputs as informed suggestions rather than definitive judgments.
\item \textbf{Ambiguity in field definitions}: Despite detailed prompts, some checklist fields remain inherently ambiguous. Distinguishing between \textsf{No} (information not reported) and \textsf{NA} (not applicable) often requires nuanced domain expertise and contextual judgment that is difficult to fully specify in prompts. Similarly, determining what constitutes sufficient reporting, for instance, whether implicit information inferred from experimental descriptions counts as \emph{reported}, involves subjective interpretation where reasonable assessors may disagree.
\item \textbf{Human assessment variability}: Not all disagreements between RECAP and human assessment necessarily indicate RECAP errors. Human annotation is itself subject to variability, bias, and occasional errors. Some apparent RECAP \emph{errors} may reflect legitimate alternative interpretations or cases where the human assessment was overly strict or lenient.
\item \textbf{Generalizability}: Our evaluation is limited to 168 papers from \gls{gecco}, focusing on \gls{ec} research. While we expect the approach to transfer to other computational domains with similar reproducibility concerns, we cannot claim generalizability without further validation. Conferences in other fields may have different reporting conventions, artifact expectations, or terminology that would require prompt adaptation.
\end{itemize}

RECAP is intended as a complement to human review, not a replacement. A Cohen's $\kappa$ of 0.67, while indicating substantial agreement, still reflects meaningful disagreement on approximately 33\% of predictions. We envision RECAP as a first-pass tool that can flag papers for further review, identify potential reproducibility gaps, and reduce reviewer workload, but final assessments should involve human judgment, particularly for borderline cases.

\subsection{Practical Recommendations for Improving Reproducibility Reporting}
\label{sec:discussion:recommendations}

The limitations discussed above point out recurring patterns that affect reproducibility-relevant information. In response, we now formulate a set of practical recommendations aimed at improving the exposition, location, and maintenance of such information.

\begin{itemize}[topsep=0.5em,leftmargin=1em,labelwidth=*,align=left]

\item \textbf{Standardized and visible artifact access}: 
If artifacts are provided, their access point should be clearly exposed and standardized, both within the paper (e.g., in a dedicated section) and on the proceedings website alongside the supplementary material. Artifacts should also include a minimal, curated description of their contents (e.g., code, data, scripts, configuration files) and execution requirements, with illustrative usage examples.

\item \textbf{Checklist-driven reproducibility workflows}: 
Authors should be encouraged to complete a reproducibility checklist at submission time as a self-audit mechanism, supported by structured templates for reporting experimental pipelines (instances, parameters, tuning protocols, evaluation measures, and environments). Reviewers, in turn, should be supported with lightweight checklist-guided prompts to enable systematic and consistent assessment of experimental reporting and artifact availability.
\end{itemize}

\section{Conclusions}
\label{sec:conclusions}

We presented a ten-year longitudinal study of reproducibility reporting in \includedPapers{} full papers from the \gls{ecom} track of \gls{gecco} (2016--2025). Using a structured checklist, we quantified how often core reproducibility signals are reported and how these practices evolve over time. We found a gradual, yet not significant, improvement in paper-level completeness, but also persistent gaps in operational details that are needed to rerun experiments.

Artifact provision remains limited. Only \somethingAvailablePerc{} of the papers provide any additional material beyond the manuscript. When artifacts are provided, they are often incomplete, weakly documented, or not persistently hosted, which reduces their long-term value.

We introduced and evaluated \emph{RECAP}, an \gls{llm}-based pipeline that extracts reproducibility signals using the same checklist as the manual protocol. The agreement with the human assessment (\cohensKappa{}) suggests that automated assessment can support scalable screening and auditing, thus helping both authors and conference organizers. 

Future work could extend the analysis to additional \gls{gecco} tracks and other conferences to assess the generality of the observed patterns. A finer-grained evaluation at the experiment level is required for papers that report multiple experimental pipelines within a single contribution. On the automation side, RECAP can be improved by more robust repository handling, better processing of large or heterogeneous artifacts, and more precise field definitions. Another possible direction is a scoring system that aggregates field-level assessments into an overall reproducibility score, providing researchers/reviewers with a quantitative measure of a paper's reproducibility.
Finally, integrating automated assessments into a human-in-the-loop review workflow may help reviewers focus their effort on papers with higher reproducibility risk, improving consistency without increasing review burden.
\\\\
\textbf{Additional Material.} This work is supported by supplementary material provided in PDF format. The dataset and the code of our analysis are available at \url{https://github.com/francesdaros/ec-reproducibility-analysis} (commit: \textsf{03bafde}) whereas RECAP is available at \url{https://github.com/TarikZ03/RECAP} (commit: \textsf{a85e181}). %Both repositories have been anonymized for review purposes. Both repositories will be mirrored on \textsf{Zenodo} upon paper acceptance.

\begin{acks}

We thank Prof.\ R.\ Bellio for the discussions on statistical testing.

The authors acknowledge the use of \glspl{llm} to support language editing, including grammar correction and rephrasing.
Following its use, all text was critically reviewed, verified, and revised by the authors

\end{acks}

\bibliographystyle{ACM-Reference-Format}
\bibliography{bibliography}

\appendix

\section{Material Collection Pipeline}
\label{sec:material-collection-pipeline}

We provide a detailed account of the data collection and selection protocol used to construct the corpus of papers analyzed in this study. While the main paper reports a concise summary of the procedure, here we describe the full retrieval process, inclusion and exclusion criteria, and the resulting dataset composition.

All candidate papers were retrieved from the \gls{acm} Digital Library, which hosts the official proceedings of \gls{gecco}.

We set the following inclusions and exclusion criteria:
\begin{enumerate}[topsep=0.5em,leftmargin=1em,labelwidth=*,align=left]
    \item A paper is included if it was published in the \gls{ecom} track of \gls{gecco};
    \item A paper is included if it was a full paper, so a paper is excluded if it is a poster, late-breaking abstract, tutorial, or workshop contribution;
    \item A paper is included if it appeared in one of the \gls{gecco} editions from 2016 to 2025 (extremes included).
     \item\label{item:coi} A paper is excluded if it is authored or co-authored by authors of the present study.
\end{enumerate}

The \gls{gecco} proceedings website facilitated reliable retrieval of the required materials. 
\gls{gecco} publishes all full research papers in the main conference proceedings and reserves posters and other contributions for the Companion Proceedings. 
This separation allowed us to identify full papers unambiguously. Moreover, the proceedings  website supports direct filtering by year and by track, enabling us to extract all 
contributions assigned to the \gls{ecom} track for each edition between 2016 and 2025. 
For each entry, the ACM Digital Library provides the PDF, bibliographic metadata, and any associated supplemental materials, which we downloaded for inclusion in our dataset. On the contrary, external artifacts (e.g., code repositories, datasets, benchmark archives) referenced in the papers are collected and evaluated as outlined in the Methodology section of the main manuscript.

\begin{figure}[ht]
    \centering
    \includegraphics[width=\linewidth]{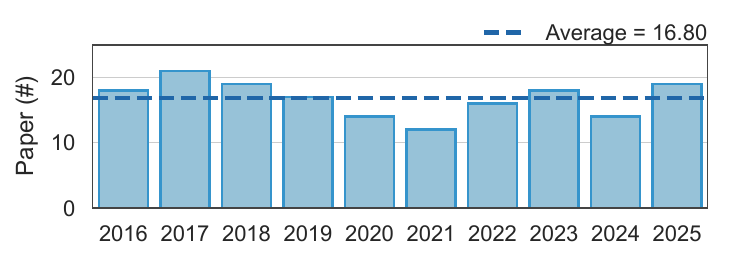}
    \Description{Number of included papers per year. The y-axis reports the number of papers. The x-axis reports the year. The average number of papers per year (16.9) is indicated with a dashed line.}
    \caption{Number of included papers per year.}
    \label{fig:papers-per-year}
\end{figure}

After downloading all materials, we apply the remaining inclusion/exclusion criteria (i.e., criterion~\ref{item:coi}). 
The final dataset comprises \includedPapers{} full papers. The distribution of papers across the years 2016–-2025 is reported in \Cref{fig:papers-per-year}.

\section{Reproducibility Checklist}
\label{sec:reproducibility-checklist}

\Cref{tab:checklist} reports the checklist, structured by dimension and indicating, for each aspect, the corresponding possible values.

\begin{table*}[ht]
    \centering
    \caption{Complete reproducibility checklist. The abbreviations are as follows. \textsf{CO: Combinatorial Optimization; COP: Combinatorial Optimization Problem; Y: Yes; N: No; NA: Not Applicable.} The reference to the ACM badge system is as per \citet{acm-badge-guidelines}}
    \label{tab:checklist}
    \small
    \begin{tabular}{lp{4.5cm}p{4.5cm}p{4.9cm}}
    \toprule
    
    Dimension & Aspect & Values & Notes \\
    
    \midrule
    
    Paper metadata & Author affiliations &	\textsf{Academic, Industry, Mixed, Not stated} & Available in the paper header.\\
    & Type of paper	& \textsf{Research article, Literature review, Tool paper, Position paper, Theory paper} & \\
    & Best paper nomination	& \textsf{Y, N} & Available in the \gls{gecco} website or in the proceesings. \\
    & Best paper award & \textsf{Y, N} & Available in the \gls{gecco} website or in the proceesings. \\
    & Supplementary material & \textsf{Y, N} & Available in the proceedings website. \\
    
    \midrule
    
    Methodology	& Pseudocode  & \textsf{Y, N, NA} & \\
    & COP natural language description & \textsf{Y, N, NA} & \\
    & COP mathematical formulation & \textsf{Y, N, NA} & \\

    \midrule
    
    Experimental setup & Hardware / machine description & \textsf{Y, N, NA} \\
    & Tuning/training instances specified & \textsf{Y, N, NA} \\
    & Testing/validation  instances specified & \textsf{Y, N, NA} \\
    & Baselines documented & \textsf{Y, N, NA}\\
    & Parameter setting & \textsf{Automated, Manual, NA} \\
    & Parameter tuning tool & \textsf{Y, N, NA} & Typically associated with \textsf{Automated setting: Automated}. \\
    & Parameter ranges & \textsf{Y, N, NA} \\
    & Tuning budget & \textsf{Y, N, NA} \\
    & Final parameter values  & \textsf{Y, N, NA} \\
    & Tuning/training random seed & \textsf{Y, N, NA} \\
    & Testing/validation random seed & \textsf{Y, N, NA} \\
    & Number of runs & \textsf{Y, N, NA} \\
    \midrule
    Results reporting & Evaluation metric defined & \textsf{Y, N, NA} \\
    & Aggregated results & \textsf{Y, N, NA} \\
    & Aggregated timing info & \textsf{Y, N, NA} \\
    & Statistical test & \textsf{Y, N, NA} \\
    & Significance level & \textsf{Y, N, NA} \\
    & Multiple-testing correction & \textsf{Y, N, NA} \\
    
    \midrule

    Artifact & Artifact provided & \textsf{Y, N, NA} \\
    & Artifact link accessible & \textsf{Y, N, NA} \\
    & Persistent artifact link & \textsf{Y, N, NA} & ACM badge \textsf{Artifacts Available v1.1}\\
    & Algorithm source code & \textsf{Y, N, NA} \\
    & Data processing code & \textsf{Y, N, NA} \\
    & Analysis / plotting code & \textsf{Y, N, NA} \\
    & Raw objective values & \textsf{Y, N, NA} \\
    & Raw solution data & \textsf{Y, N, NA} & Explicit reference to the CO field.\\
    & Raw execution times & \textsf{Y, N, NA} \\
    
    & Artifact documentation & \textsf{Y, N, NA} & \multirow{4}{*}{ACM badge \textsf{Artifacts Evaluated – Functional v1.1}} \\
    & Artifact consistency & \textsf{Y, N, NA} \\
    & Artifact completeness & \textsf{Y, N, NA} \\
    & Artifact executable & \textsf{Y, N, NA} \\
    
    & Artifact reusable & \textsf{Y, N, NA} & ACM badge \textsf{Artifacts Evaluated – Reusable v1.1}\\
    
    & Artifact reproducible & \textsf{Y, N, NA} \\
    \bottomrule
    \end{tabular}
\end{table*}

\section{Metrics Definition}
\label{sec:metrics-definition}

In this section, we report the formal definition of the metrics used in the results section of the paper.

\begin{description}
    \item[Completeness.] 
    Paper-level measure. For a given paper, the ratio between the number of checklist items marked as \textsf{Yes} and the total number of applicable items, i.e., items marked as \textsf{Yes} or \textsf{No} (excluding \textsf{NA}).

    \item[Reporting rate.] 
    Item-level measure. For a given checklist item, the ratio between the number of papers for which the item is marked as \textsf{Yes} and the total number of papers for which the item is applicable, i.e., papers marked as \textsf{Yes} or \textsf{No}.
    
    \item[Accuracy.] 
    Used to compare RECAP and the manual assessment. 
    The ratio between the number of correctly predicted checklist items and the total number of comparable items, i.e., items where both the automated prediction and the manual annotation are present.
    Formally:
    \[\text{Accuracy} = \frac{\text{Number of correct predictions}}{\text{Total number of comparable items}}\]
    
    \item[Cohen's $\kappa$ (Kappa).] 
    Used to compare RECAP and the manual assessment.  
    A measure of inter-rater agreement that accounts for agreement occurring by chance. It is defined as:
    \[\kappa = \frac{P_o - P_e}{1 - P_e}\]
    where $P_o$ is the observed agreement (i.e., accuracy) and $P_e$ is the expected agreement by chance, calculated as:
    \[P_e = \sum_{c \in C} P(\text{reviewer}_1 = c) \cdot P(\text{reviewer}_2 = c)\]
    where $C$ is the set of possible categories. Values range from $-1$ (complete disagreement) to $1$ (perfect agreement), with $0$ indicating agreement no better than chance. 
\end{description}

\section{RECAP Paper Exclusion List}
\label{sec:recap-paper-exclusion-list}

Of the 168 papers in our dataset, 13 could not be processed by RECAP due to context window limitations. Our approach loads the full paper text along with repository contents (truncated to 1{,}000 tokens per file) into the \glspl{llm} context. For papers with extensive codebases or numerous repository files, this exceeds the model's token limit per message.

Therefore, the following papers are excluded from the automated assessment \cite{10.1145/3071178.3071238,10.1145/3071178.3071235,10.1145/3205455.3205620,10.1145/3321707.3321753,10.1145/3321707.3321845,10.1145/3377930.3390224,10.1145/3377930.3389830,10.1145/3377930.3390146,10.1145/3512290.3528750,10.1145/3583131.3591054,10.1145/3583131.3590504,10.1145/3638529.3654042,10.1145/3638529.3654028}. Nevertheless, they are included in the manual assessment. % Specifically, these papers were included in the human assessment but excluded from the RECAP result analysis and comparison to the human assessment. 

\section{Experiment System Description}
\label{sec:experimental-setup}

Experiments are conducted on a machine running Ubuntu 24.04.3 LTS with an Intel Core Ultra 5 125U processor (8 cores) and 16 GB of RAM. The packages used are described in the artifact repository of RECAP (linked in the main manuscript) along with instructions of how to setup the pipeline yourself.

For code execution, we use sandboxed Docker containers running Python 3.11-slim with pre-installed scientific libraries (NumPy, Pandas, Matplotlib, SciPy, PyTorch CPU), compilers (GCC, G++, CMake), Java OpenJDK, R, and a LaTeX distribution.

The analysis of the results (both for the manual-based assessment and the comparison between this and RECAP) is conducted using Python and major libraries for vizualiations (Matplotlib and Seaborn). The code, the datasets, and the requirements for the setup are linked in the main manuscript.

\section{Additional RECAP analyses}
\label{sec:additional-recap-analysis}

%Figures \ref{fig:RECAP_confusion_matrix} -\ref{fig:RECAP_error_pattern_heatmap} 

\begin{figure}[ht]
    \centering
    \includegraphics[width=\columnwidth,trim=1mm 1mm 1mm 6mm,clip]{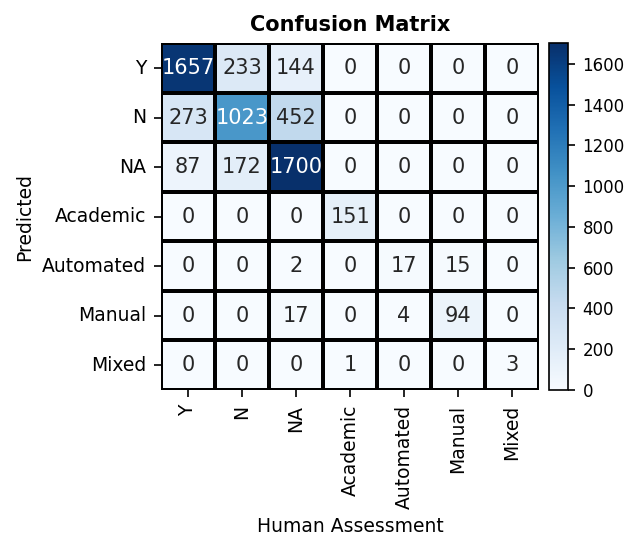}
    \caption{Confusion matrix comparing RECAP predictions (rows) against human assessment (columns) across response categories.}
    \label{fig:RECAP_confusion_matrix}
    \Description{Confusion matrix comparing RECAP predictions (rows) against human assessment (columns) across response categories.}
\end{figure}

\begin{figure}[ht]
    \centering
    \includegraphics[width=\columnwidth]{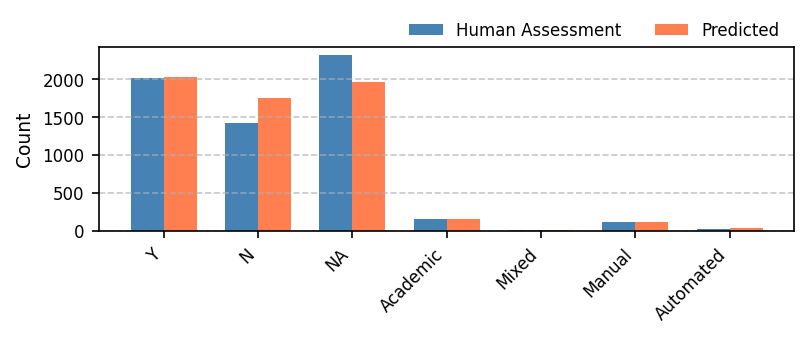}
    \caption{Distribution of responses in human assessment versus RECAP predictions. Both exhibit similar patterns, with \textsf{Y} and \textsf{NA} being most frequent.}
    \label{fig:RECAP_class_distribution}
    \Description{Distribution of responses in human assessment versus RECAP predictions. Both exhibit similar patterns, with Yes and NA being most frequent.}
\end{figure}

\begin{figure}[ht]
    \centering
    \includegraphics[width=\columnwidth]{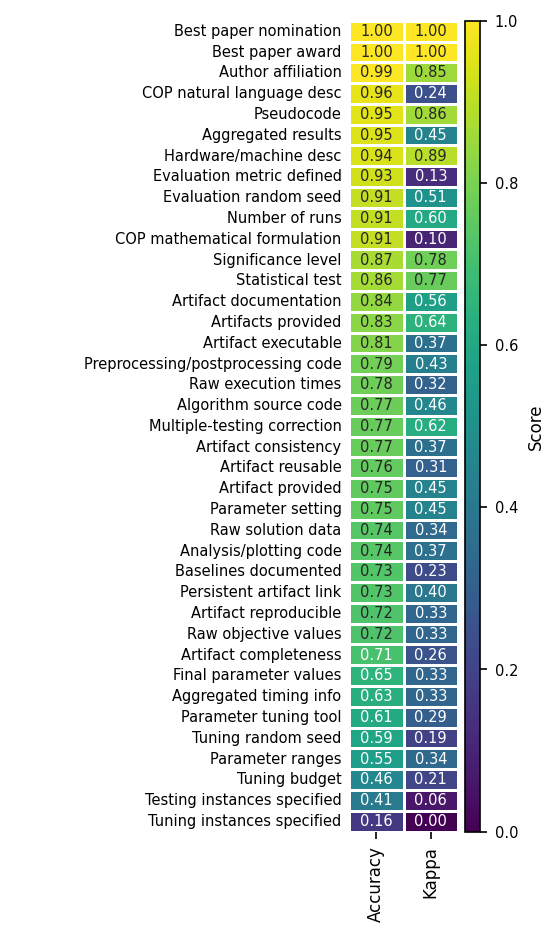}
    \caption{Per-field accuracy and Cohen's $\kappa$ (Kappa), sorted by accuracy. Fields related to tuning and training/testing instances show the lowest agreement.}
    \label{fig:RECAP_field_heatmap}
    \Description{Per-field accuracy and Cohen's Kappa, sorted by accuracy. Fields related to tuning and training/testing instances show the lowest agreement.}
\end{figure}

\begin{figure}[ht]
    \centering
    \includegraphics[width=\columnwidth]{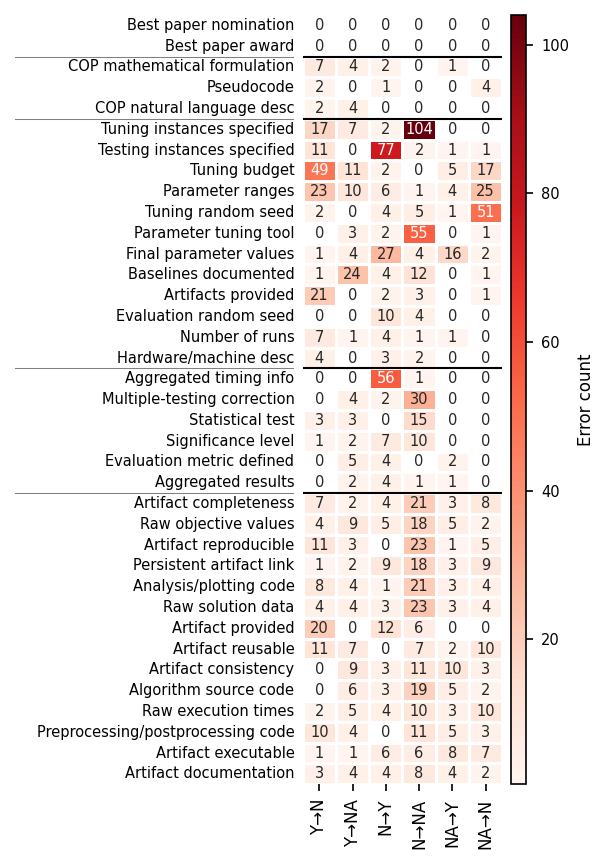}
    \caption{Error patterns by field showing confusion types (Predicted $\rightarrow$ Human Assessment). Darker cells indicate more frequent misclassifications}
    \Description{Error patterns by field showing confusion types (Predicted to Human Assessment). Darker cells indicate more frequent misclassifications}
    \label{fig:RECAP_error_pattern_heatmap}
\end{figure}

\Cref{fig:RECAP_confusion_matrix,fig:RECAP_error_pattern_heatmap,fig:RECAP_class_distribution,fig:RECAP_field_heatmap} provide a detailed breakdown of RECAP's prediction behavior.

\Cref{fig:RECAP_confusion_matrix} reports a confusion matrix comparing the RECAP assessment with the manual one. 
RECAP correctly predicts the majority of \textsf{Y}, \textsf{N}, and \textsf{NA} responses, with most predictions falling along the diagonal. 
It is worth noting that some disagreements may stem from variability in human assessment rather than RECAP errors, as annotators may differ in their strictness when interpreting edge cases.

The class distribution comparison confirms that RECAP's response patterns closely mirror human assessment (\Cref{fig:RECAP_class_distribution}). Both distributions show \textsf{Y} and \textsf{NA} as the most frequent responses, with RECAP slightly over-predicting \textsf{N} and under-predicting \textsf{NA}. This supports the argument that different annotators differ in their interpretation of this kind of material.

The field-level heatmap from \Cref{fig:RECAP_field_heatmap} gives insight into the performance of RECAP across fields. Fields achieving high accuracy (>0.90) and $\kappa$ (>0.60) include \textsf{Best paper award, Best paper nomination, Author affiliation, Pseudocode, Number of runs, Hardware/machine description}, to name a few. We believe that this because these fields are typically well-defined and unambiguous in papers.

Conversely, fields with low accuracy (<0.50) include \textsf{Tuning/training instances} (0.16), \textsf{Testing/validation Instances} (0.39), and \textsf{Tuning budget}(0.46). This discrepancy primarily stems from systematic differences between how RECAP \emph{interpret} textual descriptions and the stricter criteria adopted in the manual assessment. For instance, considering the instances, RECAP tends to classify high-level or qualitative statements (e.g., describing proportions of instances or generic random splits) as sufficient evidence of reporting, whereas the manual assessors look for explicit operational details, such as the precise list of instances, the random seed used for splitting, or direct access to generated instances, to consider an item reproducible. A similar mismatch arises when papers reference instance generators or derived benchmark sets without providing either the generator implementation or the resulting instances. In these cases, experimental choices are conceptually described but remain impractical to reproduce. As a consequence, the automated system overestimates compliance for these items, leading to lower agreement with the manual assessment.

The error heatmap in \Cref{fig:RECAP_error_pattern_heatmap} identifies specific confusion patterns per field. Several notable patterns emerge:

\begin{itemize}[topsep=0.5em,leftmargin=1em,labelwidth=*,align=left]
    \item \textbf{N$\rightarrow$NA and NA$\rightarrow$N confusion}: Fields like \textsf{Tuning/training instances} (104 errors), \textsf{Parameter tuning tool} (55), \textsf{Multiple-testing correction} (30) and \textsf{Tuning/trianing random seed} (51) show high confusion between \textsf{N} and \textsf{NA}. This suggests RECAP struggles to distinguish between information that is not reported and not applicable, which is a distinction that often requires domain expertise about whether a field should apply to a given paper. 
    
    \item \textbf{Y$\rightarrow$N and N$\rightarrow$Y confusion}: \textsf{Tuning budget} (49), \textsf{Testing/validation instances} (77), and \textsf{Aggregated timing info} (56) indicate regular errors where \textsf{Y} and \textsf{N} have been swapt. Apart from these fields, the error rates for all other fields are very low, and in most cases marginal. However, with these fields sometimes the information must be inferred from experimental descriptions rather than explicitly stated, whereas RECAP tries to find information that is explicitly stated.
\end{itemize}

These patterns suggest that the primary challenge does not lie in identifying \textit{presence} of information, rather in distinguishing between \emph{absence} \textsf{(N)} and \emph{inapplicability} \textsf{(NA)}, as well as between \emph{explicit} and \emph{implicit} reporting. This distinction often requires understanding the paper at a deeper level, for instance, knowing that a theoretical paper would not have tuning parameters, versus an empirical paper that simply failed to report them.

The error analysis points to several potential improvements: 
\begin{enumerate*}[label=(\roman*)]
    \item providing more explicit definitions of what constitutes each possible answer, 
    \item supplying additional metadata to help RECAP determine field applicability, and 
    \item implementing a multi-pass verification for challenging fields where initial assessments are re-evaluated with targeted follow-up queries.
\end{enumerate*}
% These refinements could substantially improve performance on the currently challenging fields.

\section{Article Checklist}
\label{sec:article-checklist}

\Cref{tab:checklist-this-paper} reports the completed checklist of this paper. 

\begin{table*}[ht]
    \centering
    \caption{Reproducibility checklist of our paper.}
    \label{tab:checklist-this-paper}
    \small
    \begin{tabular}{lp{4.5cm}p{4.5cm}p{4.9cm}}
    \toprule
    
    Dimension & Aspect & Value & Notes \\
    
    \midrule
    
    Paper metadata & Author affiliations &	-- & Anonymous author(s).\\
    & Type of paper	& Research article & \\
    & Best paper nomination	& -- & To be completed after review \\
    & Best paper award & -- & To be completed after review \\
    & Supplementary material & \textsf{Y} &  \\
    
    \midrule
    
    Methodology	& Pseudocode  & \textsf{NA} & However, flowcharts are reported to explain assessment protocols. \\
    & COP natural language description & \textsf{NA} & The paper does not describe a Combinatorial Optimization work. \\
    & COP mathematical formulation & \textsf{NA} & The paper does not describe a Combinatorial Optimization work.  \\

    \midrule
    
    Experimental setup & Hardware / machine description & \textsf{Y} & Reported in the supplementary material. Machine running Ubuntu 24.04.3 LTS with an Intel Core Ultra 5 125U processor (8 cores) and 16 GB of RAM. \\
    & Tuning/training instances specified & \textsf{NA} \\
    & Testing/validation  instances specified & \textsf{Y} &  (\emph{Item interpretation}) Selection of papers is described in Section 3.1, further motivated in the supplementary material\\
    & Baselines documented & \textsf{Y} & (\emph{Item interpretation}) Human assessment is the baseline, documented in Section 3.3.  \\
    & Parameter setting & \textsf{NA} \\
    & Parameter tuning tool & \textsf{NA}  \\
    & Parameter ranges & \textsf{NA} \\
    & Tuning budget & \textsf{NA} \\
    & Final parameter values  & \textsf{NA} \\
    & Tuning/training random seed & \textsf{NA} \\
    & Testing/validation random seed & \textsf{NA} \\
    & Number of runs & \textsf{Y} & 1 per paper \\
    \midrule
    Results reporting & Evaluation metric defined & \textsf{Y} & In the supplementary material \\
    & Aggregated results & \textsf{Y} & Section 4 of the main paper, further analysis reported in the supplementary material \\
    & Aggregated timing info & \textsf{NA} \\
    & Statistical test & \textsf{Y} & Section 4 of the main paper, specifically in Section 4.1 \\
    & Significance level & \textsf{Y} & Section 4, both test set at $\alpha=0.05$ \\
    & Multiple-testing correction & \textsf{NA} & Not required. \\
    
    \midrule

    Artifact & Artifact provided & \textsf{Y} & Thrugh anonymized link\\
    & Artifact link accessible & \textsf{Y} \\
    & Persistent artifact link & \textsf{N} & But the repositories will be mirrored to \textsf{Zenodo} upon paper acceptance \\
    & Algorithm source code & \textsf{Y} \\
    & Data processing code & \textsf{Y} \\
    & Analysis / plotting code & \textsf{Y} \\
    & Raw objective values & \textsf{Y} \\
    & Raw solution data & \textsf{NA} &\\
    & Raw execution times & \textsf{NA} \\
    
    & Artifact documentation & \textsf{Y} & \\
    & Artifact consistency & \textsf{Y} \\
    & Artifact completeness & \textsf{Y} \\
    & Artifact executable & \textsf{Y} \\
    
    & Artifact reusable & \textsf{Y} & \\
    
    & Artifact reproducible & \textsf{Y} \\
    \bottomrule
    \end{tabular}
\end{table*}

\end{document}